\documentclass[journal]{IEEEtran}
\ifCLASSINFOpdf
  \usepackage[pdftex]{graphicx}
\else
\fi
\usepackage{amsmath}
\usepackage{amssymb}
\usepackage{array}
\usepackage{booktabs}

  \usepackage[caption=false,font=footnotesize]{subfig}

\usepackage{graphicx}
\usepackage{multirow}

\begin{document}
\title{Temporal Action Localization using Long Short-Term Dependency}

\author{Yuan~Zhou,~\IEEEmembership{Senior Member,~IEEE,}
        Hongru~Li
        and~Sun-Yuan~Kung,~\IEEEmembership{Life~Fellow,~IEEE}% <-this % stops a space
\thanks{Yuan Zhou and Hongru Li are with the School of Electrical
and Information Engineering, Tianjin University, Tianjin 300072, China. Corresponding author: Yuan Zhou (e-mail: zhouyuan@tju.edu.cn).}
% <-this % stops a space
\thanks{Sun-Yuan Kung, is with the Department of Electrical Engineering, Princeton
University, Princeton, NJ 08540, USA. (e-mail: kung@princeton.edu)}% <-this % stops a space
}


\markboth{IEEE Transactions on Multimedia,~Vol.~XXX, No.~XXX}%
{Shell \MakeLowercase{\textit{et al.}}: Bare Demo of IEEEtran.cls for IEEE Journals}



\maketitle

\begin{abstract}
Temporal action localization in untrimmed videos is an important but difficult task. Difficulties are encountered in the application of existing methods when modeling temporal structures of videos. In the present study, we developed a novel method, referred to as Gemini Network, for effective modeling of temporal structures and achieving high-performance temporal action localization. The significant improvements afforded by the proposed method are attributable to three major factors. First, the developed network utilizes two subnets for effective modeling of temporal structures. Second, three parallel feature extraction pipelines are used to prevent interference between the extractions of different stage features. Third, the proposed method utilizes auxiliary supervision, with the auxiliary classifier losses affording additional constraints for improving the modeling capability of the network. As a demonstration of its effectiveness, the Gemini Network was used to achieve state-of-the-art temporal action localization performance on two challenging datasets, namely, THUMOS14 and ActivityNet.
\end{abstract}


% \fi
\IEEEpeerreviewmaketitle

\section{Introduction}

\IEEEPARstart{O}{wing} to the explosive growth of video data, video content analysis has attracted significant attention in both industry and academia in recent years. An important aspect of video content analysis is action recognition, which is usually used to classify manually trimmed video clips. Substantial progress has been achieved in the development of the technology \cite{8454294,sun2017lattice,shi2017learning,wu2018compressed}. However, most videos in the real world are untrimmed and may contain multiple action instances with irrelevant background scenes or activities. This has stimulated research interest in another related aspect of video analysis, namely, temporal action localization. Temporal action localization is a challenging task that requires not only the determination of the category of video clips, but also identification of the temporal boundaries (starting and ending time points) of the action instances in the untrimmed videos. Temporal action localization is crucial to several important subjects of video technology, ranging from direct applications such as the extraction of highlights from sports videos to higher-level tasks such as automatic video captioning.

The aim of temporal action localization is to determine the time boundaries of actions in a video. So temporal action localization requires modeling of not only spatial structures, but also temporal structures. For modeling spatial structures, convolutional neural network (CNN) is widely used owing to its high power for spatial object detection \cite{shou2017cdc,lin2017single,chao2018rethinking,paul2018w}. However, there is no common framework in temporal structure modeling, which is also important for temporal action localization. Several different frameworks are used to model temporal structures, such as the Convolutional-De-Convolutional (CDC) network \cite{shou2017cdc}, which utilizes 3D ConvNet (C3D) for temporal structure modelling \cite{tran2015learning}. However, the architecture of C3D limits its capability to capture long-range temporal dependencies. Some researchers proposed the Single Shot Action Detector (SSAD) \cite{lin2017single} to model temporal structures, which utilizes two-stream ConvNet \cite{simonyan2014two}. However, the spatio-temporal features, which are extracted from each short temporal segment, are simply fused to represent the entire proposal in a fully connected (FC) layer. This results in the long-range temporal dependencies not being properly captured. Some researchers have also used recurrent neural network (RNN) \cite{buch2017sst,buch2017end,alwassel2018action}. However, LSTM could only sharply distinguish a nearby context from a distant history and has limited capability for capturing long-term dependencies.

\begin{figure*}[t]
\begin{center}
\includegraphics[width=16.2cm]{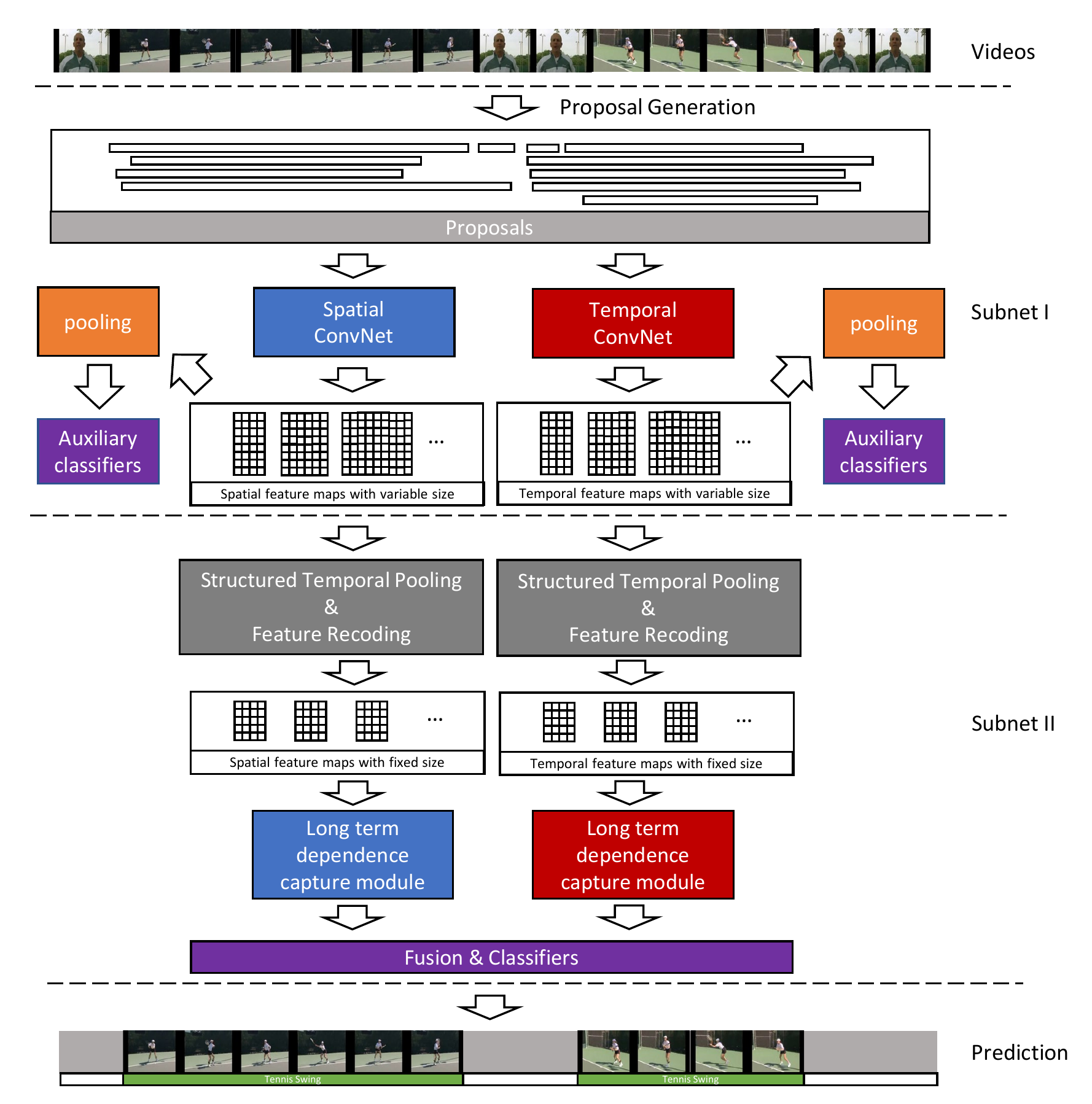}
\end{center}
   \caption{Illustration of the overall system. From top to bottom is the execution procedure. Given an input video, we first generate temporal proposals by using temporal segment networks and the temporal actionness grouping algorithm. We then extract spatiotemporal features by the proposed Gemini Network which contains two subnets:  subnet I models the short-range temporal structures of video data, while subnet II models the long-range ones. Finally, the extracted spatiotemporal features are fed into classifiers to generate predictions.}
\label{img1}
\end{figure*}

All the above methods have limitations. They only use a single deep neural network to model the temporal structure, whereas temporal structures have two types of temporal dependencies, namely, short-range and long-range dependencies. Hence, the methods cannot be used to accurately capture both types of dependencies by a single deep neural network. It is well recognized that, for a specific task, an application-specific framework performs better than universal frameworks. Likewise, to overcome the above limitations, our idea is to construct two application-specific networks rather than a single universal network for temporal dependencies extraction.

In this paper, we propose a new network, termed as Gemini Network, which utilizes long and short-term dependencies respectively in two subnets (see Figure \ref{img1}). 1) Subnet I contains a two-stream ConvNet, with the spatial-stream ConvNet accepting the original video frames as input and extracting spatial features from them, and the temporal-stream ConvNet accepting the optical flow stack as input and extracting short-range temporal features from them. 2) Subnet II models long-range temporal structures. In Subnet II, parallel temporal structure pipelines are used to extract features from the action instance and its time context. Finally, the classifiers produce the predictions of the proposals. The main contributions of this study are threefold:
\begin{enumerate}
\item To the best of our knowledge, this work is the first to apply respective modeling for different temporal features. We divide temporal structures into two types: short-range and long-range dependencies, and respectively capture them in two specific subnets. By elaborately controlling the temporal receptive fields of two subnets, each subnet can be more specific in modeling different types of temporal structures, resulting in higher performance.

\item We introduce auxiliary supervision for accurate modeling of the temporal structures. More specially, several additional auxiliary loss functions were added to Subnet I to enable the acquisition of additional regularizations and increase the strength of the gradient flow that gets propagated back.

\item We conduct extensive experiments on two benchmarks for temporal action localization, namely THUMOS'14 \cite{THUMOS14} and ActivityNet \cite{Heilbron2015ActivityNet}. The results demonstrate that the proposed approach can achieve state-of-the-art detection performance compared with other existing methods.
\end{enumerate}

\section{Related Work}

\subsection{Action Recognition}
Action recognition is an important aspect of video content analysis and has been extensively studied over the last few years \cite{Zhen2017Supervised,Hou2017Content,Shi2017Sequential,sun2018optical,Wang2018Two,Roy2018Unsupervised}. Earlier methods are mostly based on hand-crafted visual features \cite{laptev2005space,wang2011action}. Among these methods, Improved Dense Trajectory Feature (iDTF) method \cite{wang2011action}, which utilizes HOG, HOF, and MBH features extracted along dense trajectories, has been predominantly used. However, the use of convolutional neural networks (CNNs) has been recently shown to significantly improve the performance of object detection. CNNs were first applied to action recognition in \cite{Karpathy2014Large}, although limited performance was achieved because they only captured spatial information. The use of two-stream ConvNets was subsequently proposed for utilization of the temporal information of video sequences \cite{simonyan2014two}. In this method, both RGB and optical flow are used as inputs for the model motion. The idea is further developed in temporal segment networks \cite{8454294}, which affords remarkable performance. Later, C3D architecture \cite{tran2015learning} was proposed for direct extraction of the spatio-temporal features of high-level semantics from raw videos. Effort has also been made to model long-range temporal dependencies using temporal convolution or recurrent neural networks (RNNs) \cite{donahue2015long,sun2017lattice}. However, these methods assume well-trimmed videos with the action of interest lasting for nearly the entire duration. There is thus no need to consider the localization of the action instances.

\subsection{Temporal Action Localization}
In early works in this area, sliding windows were used to generate proposals, with a focus on the design of hand-crafted feature representations for category classification \cite{oneata2013action,tang2013combining}. However, the performance of the employed methods was limited by the hand-crafted feature representations. Recently, deep networks were applied to action localization to achieve improved performance \cite{shou2016temporal,8579103,Liu2018Weakly,heilbron2018annotate,alwassel2018action,lin2018bsn}. In S-CNN  \cite{shou2016temporal}, multi-stage CNN that utilizes 3D-ConvNet (C3D) \cite{tran2015learning} is used to simultaneously capture the spatial and temporal features. However, the architecture of C3D limits its capability to capture long-range temporal dependencies. Inspired by Faster R-CNN\cite{ren2015faster}, researchers \cite{xu2017r} developed an R-C3D architecture that uses 3D-ConvNet to extract features, and pooling of the 3D region of interest (RoI) to fix the size of the feature maps. However, the 3D convolutional feature extractor limits the capability of this method for modeling temporal dependencies. The pooling of the 3D region of interest (RoI) causes the classifiers to lose some important features, resulting in unsatisfactory performance.

Just as an image classification network can be used for image object detection, action recognition models can also be used for temporal action detection for feature extraction. In addition, the two-stream ConvNet architecture \cite{simonyan2014two} has been proved to be effective for action recognition \cite{8454294}. It is used in SSAD \cite{lin2017single} to extract spatio-temporal features for proposal classification. However, the spatio-temporal features, which are extracted from each short temporal segment, are simply fused to represent the entire proposal in a fully connected (FC) layer. This results in the long-range temporal dependencies not being properly captured.

Some new methods were recently developed \cite{buch2017sst,buch2017end,Singh2016A} for modeling the temporal evolution of activities using long-short-term memory (LSTM) networks, with the prediction of an activity label performed at each time step. However, \cite{khandelwal2018sharp} found that LSTM could only sharply distinguish a nearby context from a distant history. In other words, LSTM has limited capability for capturing long-term dependencies. Hence, CNN was preferentially used in this study to model the temporal dependencies of an arbitrary proposal granularity, with significantly higher accuracy achieved.

\section{Architecture of Gemini Network}

\begin{figure*}[t]
\begin{center}
% \fbox{\rule{0pt}{2in} \rule{.9\linewidth}{0pt}}
\includegraphics[width=15cm]{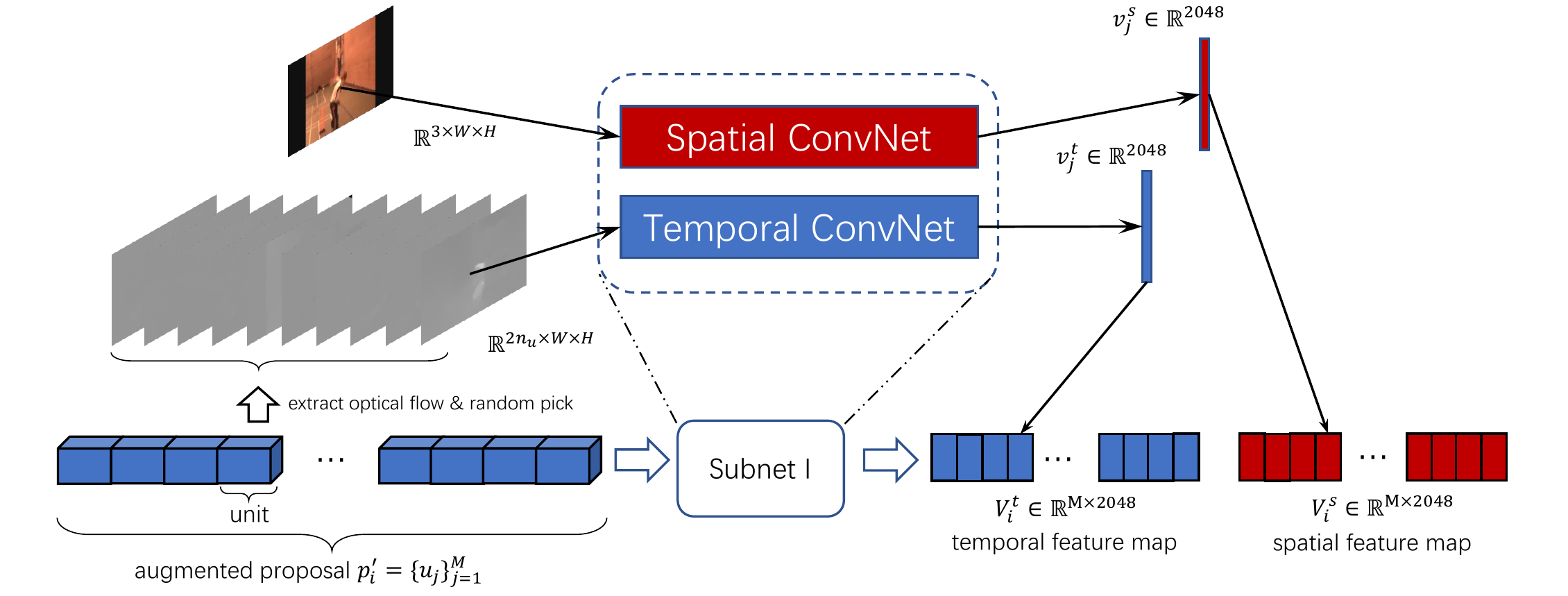}
\end{center}
   \caption{Details of Subnet I. The video units are the basic processing units. An RGB image is randomly picked from a unit and mapped to a spatial feature vector. An optical flow stack is extracted from a unit and mapped to a temporal feature vector. The spatial and temporal feature vectors are respectively concatenated to form spatial and temporal feature maps.}
\label{img3}
\end{figure*}

The proposed Gemini Network is intended for action localization in continuous video streams. As illustrated in Fig. \ref{img1}, the network consists of two components, namely, Subnet I and Subnet II. Subnet I is used for the extraction of spatial features and to capture the short-range temporal dependencies, while Subnet II is used to capture long-range temporal dependencies. The proposals are generated and augmented to contain more context information, then fed into the network¡¯s Subnet I. Then the features generated by Subnet I are subsequently fed into Subnet II to model long-range temporal structures.

\subsection{Subnet I: Extracting Spatial and Short-range Temporal Dependency}\label{sec3.2}

When applied to an action recognition task, the two-stream architecture \cite{simonyan2014two} exhibits its ability to learn video representation for the capture of temporal dependencies. Accordingly, we built Subnet I on a two-stream architecture containing a spatial stream ConvNet and a temporal stream ConvNet. Inception \cite{szegedy2015going} was specifically used to build each stream.

Subnet I is illustrated in Fig. \ref{img3}. For the spatial stream ConvNet, the input is a single RGB image randomly picked from a unit, and the output is a vector $v_j^s\in{\mathbb{R}}^{2048}$ (here, $s$ indicates spatial). The output $v_{j}^{s}$ can be considered as the spatial feature of unit $u_{j}$. For each proposal, we concatenate the $v_{j}^{s}$ of each unit $u_{j}$ to form a feature map of augmented proposal $p_{i}'$. The spatial feature map of augmented proposal $p_{i}'$ can be represented by $V_{i}^{s}=[v_{1}^{s},v_{2}^{s},v_{3}^{s},\ldots,v_{M}^{s}]$.

For the temporal stream ConvNet, the input is a stack of consecutive optical flow fields obtained from a unit. An optical flow can be considered as a set of displacement vector fields between consecutive frames $k$ and $k+1$. Let $O_m^x$ and $O_m^y$ represent the horizontal and vertical channels of the optical flow field, respectively. The stack of the consecutive optical flow fields of unit $u_{j}$ can be denoted by $S_{j}=\{O_{m}^{x},O_{m}^{y}|m\in[1,n_{u}]\}$. The input $S_{j}\in\mathbb{R}^{2 n_{u}\times H\times W}$ is mapped to a feature vector $v_j^t\in{\mathbb{R}}^{2048}$ by the temporal ConvNet (here, $t$ indicates temporal). The temporal feature map of the augmented proposal $p_{i}'$ can be denoted as $V_{i}^{t}=[v_{1}^{t},v_{2}^{t},v_{3}^{t},\ldots,v_{M}^{t}]$.

From the foregoing, each augmented proposal can be represented by a spatial feature map $V_{i}^{s}=[v_{1}^{s},v_{2}^{s},v_{3}^{s},\ldots,v_{M}^{s}]$  and a temporal feature map $V_{i}^{t}=[v_{1}^{t},v_{2}^{t},v_{3}^{t},\ldots,v_{M}^{t}]$.

\subsection{Subnet II: Extracting Long-range Temporal Dependency}\label{sec3.3}

As Subnet II processes snippet-level features, it could capture long-range temporal dependencies.

Subnet II is designed to capture long-range temporal dependencies. It has three components: 1) Self-adaptive pooling, which is used to extract fixed-size feature maps for variable-length proposals. 2) Feature recoding, which is used to create a new representation of the fixed-size feature maps. 3) Capture module, which is used to capture long-range temporal dependencies of augmented proposals.

\begin{figure}[t]
\begin{center}
\includegraphics[width=8cm]{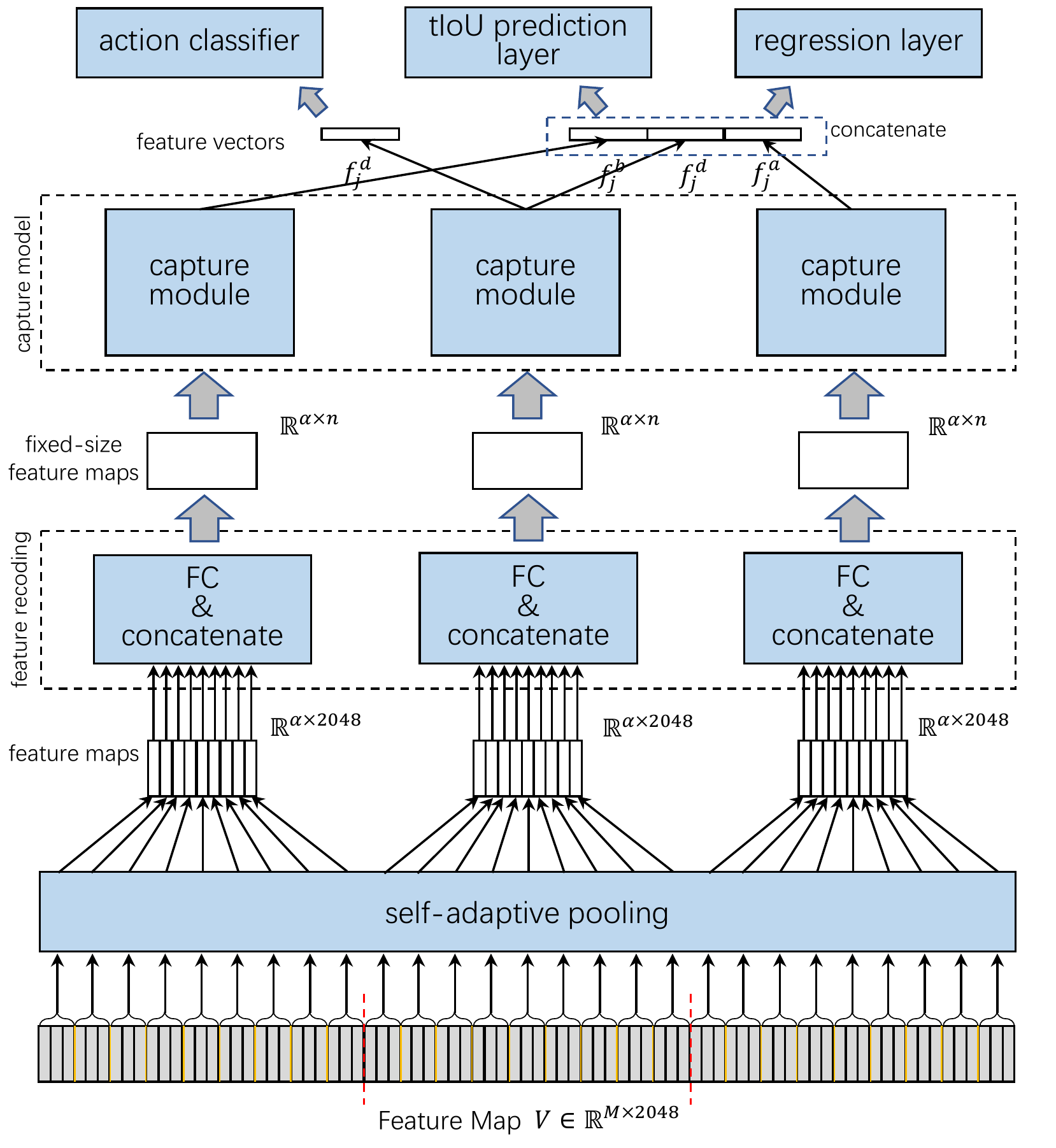}
\end{center}
   \caption{Details of Subnet II. A temporal or spatial feature map is divided into $3\alpha$ segments. Here, $\alpha$ is set to 9. FCs are used for feature recoding. Here, output dimension $n$ of FCs is 129. After self-adaptive pooling and feature recoding, three fixed-size feature maps are obtained for the three stages. The three feature maps are then mapped into three feature vectors, $f_j^b$, $f_j^d$ and $f_j^a$, respectively. Subsequently, $f_j^d$ is fed into action classifier and the concatenated vector $f_j=[f_j^b,f_j^d,f_j^a]$ is fed into tIoU prediction layer and the regression layer.}
\label{img4}
\end{figure}

The spatial feature map $V_{i}^{s}=[v_{1}^{s},v_{2}^{s},v_{3}^{s},\ldots,v_{M}^{s}]$ and temporal feature map $V_{i}^{t}=[v_{1}^{t},v_{2}^{t},v_{3}^{t},\ldots,v_{M}^{t}]$, which are obtained by Subnet I, are considered as the inputs to Subnet II, which is illustrated in Fig. \ref{img4}. The operations performed on the spatial and temporal feature maps are entirely the same. The spatial feature map is considered here as an example.

\subsubsection{Self-adaptive Pooling}

Spatial feature map $V_{i}^{s}=[v_{1}^{s},v_{2}^{s},v_{3}^{s},\ldots,v_{M}^{s}]$ of augmented proposal $p_{i}'$ is first divided into $3\alpha$ segments. The spatial feature map is rewritten as ${V_{i}^{s}}'=[{seg}_{1},{seg}_{2},{seg}_{3},\ldots,seg_{3\alpha}]$, where $seg_{j}=\{u_{j}|s\leqslant j\leqslant e,s=1+\frac{(j-1)M}{3\alpha},e=\frac{jM}{3\alpha}\}$ (here, $seg$ indicates segment). The spatial feature map is then fed into the self-adaptive pooling, which pools each segment of the vectors into one vector. For $seg_{j}$, the following vector can be obtained:
\begin{equation}
r_{j}=\frac{1}{|e-s+1|}\sum_{j=s}^{e}v_{j}.
\end{equation}
The self-adaptive pooling subsequently separates the vectors into three groups: before the action $R_{i}^{b}=[r_{1},r_{2},r_{3},\ldots,r_{\alpha}]$, during the action $R_{i}^{d}=[r_{\alpha+1},r_{\alpha+2},r_{\alpha+3},\ldots,r_{2\alpha}]$ and after the action $R_{i}^{a}=[r_{2\alpha+1},r_{2\alpha+2},r_{2\alpha+3},\ldots,r_{3\alpha}]$.

\subsubsection{Feature Recoding}

The three vector groups are separately fed into three fully connected (FC) layers to recode the features. Through the FC layers, vector $r_{j}$ can be mapped to a new vector as follows:
\begin{equation}
d_{j}=Wr_{j}+b,
\end{equation}
where $W\in\mathbb{R}^{n\times 2048}$, $b\in\mathbb{R}^{n}$, $n$ is the dimension of $d_{j}$, and $W_{i}$ and $b_{i}$ are respectively a weight matrix and its bias. We concatenate the new vectors $d_{j}$ in each of the three groups to form three fixed-size feature maps for the three stages: before the action $D_{i}^{b}=[d_{1},d_{2},d_{3},\ldots,d_{\alpha}]\in\mathbb{R}^{\alpha \times n}$, during the action $D_{i}^{d}=[d_{\alpha+1},d_{\alpha+2},d_{\alpha+3},\ldots,d_{2\alpha}]\in\mathbb{R}^{\alpha \times n}$ and after the action $D_{i}^{a}=[d_{2\alpha+1},d_{2\alpha+2},d_{2\alpha+3},\ldots,d_{3\alpha}]\in\mathbb{R}^{\alpha \times n}$.

\subsubsection{Capture Module}

We feed $D_j^b$, $D_j^d$ and $D_j^a$ into three capture models respectively, and computed the stage-level feature vectors $f_j^b$, $f_j^d$, $f_j^a$. The architecture of the capture module is a redesign of that of ResNet-18. The structure parameters are given in Table \ref{tab1}. The feature vector $f_j^d$ is fed into a classification layer for predicting the confidence score of each category. In addition, $f_j=[f_j^b,f_j^d,f_j^a]$ is fed into two FCs, one of which served as a temporal Intersection-over-Union (tIoU) layer for prediction the degree of overlap between the proposal and corresponding ground truth, and the other as a regression layer for generating refined start-end times.

\begin{table}[t]
\caption{Architectures of capture model in Subnet II. The building blocks are the same as those of ResNet-18 \cite{he2016deep}, where the blocks stacked on each other. Downsampling is performed by conv1, conv2\_a, conv3\_a, conv4\_a, and conv5\_a. Other convolutional layers keep the dimension of feature maps. Here, we show the case that number of segments $\alpha = 9$. Each dimension of the output represents the number of channels, height in pixel and width in pixel, respectively.}
\label{tab1}
\begin{center}
\resizebox{0.47\textwidth}{!}{
\begin{tabular}{ccc}
  \specialrule{0.05em}{0pt}{1.5pt}
  Name & Output size & Blocks \\
  \specialrule{0.05em}{1.5pt}{1.5pt}
  conv1 & 32$\times$65$\times$5 & 3$\times$3, 32, stride $2\times1$ \\ \specialrule{0em}{1.5pt}{1.5pt}
  conv2\_a & 64$\times$33$\times$5 & $\left[\begin{matrix}3\times3 & 64 \\ 3\times3 & 64\end{matrix}\right]$, stride $\left[\begin{matrix}2\times1 \\ 1\times1\end{matrix}\right]$ \\ \specialrule{0em}{1.5pt}{1.5pt}
  conv2\_b & 64$\times$33$\times$5 & $\left[\begin{matrix}3\times3 & 64 \\ 3\times3 & 64\end{matrix}\right]$, stride $\left[\begin{matrix}1\times1 \\ 1\times1\end{matrix}\right]$ \\ \specialrule{0em}{1.5pt}{1.5pt}
  conv3\_a & 128$\times$17$\times$5 & $\left[\begin{matrix}3\times3 & 128 \\ 3\times3 & 128\end{matrix}\right]$, stride $\left[\begin{matrix}2\times1 \\ 1\times1\end{matrix}\right]$ \\ \specialrule{0em}{1.5pt}{1.5pt}
  conv3\_b & 128$\times$17$\times$5 & $\left[\begin{matrix}3\times3 & 128 \\ 3\times3 & 128\end{matrix}\right]$, stride $\left[\begin{matrix}1\times1 \\ 1\times1\end{matrix}\right]$ \\ \specialrule{0em}{1.5pt}{1.5pt}
  conv4\_a & 256$\times$9$\times$5 & $\left[\begin{matrix}3\times3 & 256 \\ 3\times3 & 256\end{matrix}\right]$, stride $\left[\begin{matrix}2\times1 \\ 1\times1\end{matrix}\right]$ \\ \specialrule{0em}{1.5pt}{1.5pt}
  conv4\_b & 256$\times$9$\times$5 & $\left[\begin{matrix}3\times3 & 256 \\ 3\times3 & 256\end{matrix}\right]$, stride $\left[\begin{matrix}1\times1 \\ 1\times1\end{matrix}\right]$ \\ \specialrule{0em}{1.5pt}{1.5pt}
  conv5\_a & 512$\times$5$\times$3 & $\left[\begin{matrix}3\times3 & 512 \\ 3\times3 & 512\end{matrix}\right]$, stride $\left[\begin{matrix}2\times1 \\ 1\times1\end{matrix}\right]$ \\ \specialrule{0em}{1.5pt}{1.5pt}
  conv5\_b & 512$\times$5$\times$3 & $\left[\begin{matrix}3\times3 & 512 \\ 3\times3 & 512\end{matrix}\right]$, stride $\left[\begin{matrix}1\times1 \\ 1\times1\end{matrix}\right]$ \\ \specialrule{0em}{1.5pt}{1.5pt}
  avg pool & 512$\times$2$\times$2 & 3$\times$2, stride $2\times1$ \\ \specialrule{0.05em}{1.5pt}{0pt}
\end{tabular}}
\end{center}
\end{table}

\section{Training and Prediction}

\subsection{Proposal Generation}\label{sec4.1}

It is desirable for temporal proposal generation to have regions that can cover a wide range of durations while accurately matching the ground-truths. In existing methods, the proposal generation is implemented by sliding windows, which involves an extremely high computational cost. To avoid this disadvantage, we apply temporal actionness grouping (TAG) \cite{xiong2017pursuit} into the proposal generation method.

Let $T$ denote the number of frames in a video. A video is divided into $T/n_{u}$ consecutive video units, where $n_{u}$ is the frame number of a unit. The video unit is set as the basic processing unit; that is, the proposed framework considers the units as the inputs. Let $I_{k}$ denote the $\mathit{k}$-th frame. The $\mathit{i}$-th unit of the video can be denoted by $u_{i}=\{I_{k}\}_{k=n_{u}(i-1)+1}^{n_{u}\cdot i}$, where $n_{u}(i-1)+1$ is the starting frame and $n_{u}\cdot i$ is the ending frame.

Consider a given set of $N$ proposals $P=\{p_{i}=[s_{i},e_{i}]\}_{i=1}^{N}$. Each proposal has a starting point $s_{i}$ and ending point $e_{i}$. The duration of the proposal $p_{i}$ is given by $d_{i}=e_i-s_{i}+1$. To capture more context information, we tripled the span of $p_{i}$ by extending it beyond the starting and ending points, thereby obtaining the augmented proposal $p_{i}^{'}=[s_{i}^{'},e_{i}^{'}]$, where $s_{i}-s_{i}^{'}+1=e_{i}^{'}-e_{i}+1=d_{i}$. Consider a given augmented proposal $p_{i}'$ containing $M$ video units, we have $p_{i}^{'}=\{u_{j}\}_{j=1}^{M}$.

\begin{figure}[h]
\begin{center}
\includegraphics[width=8cm]{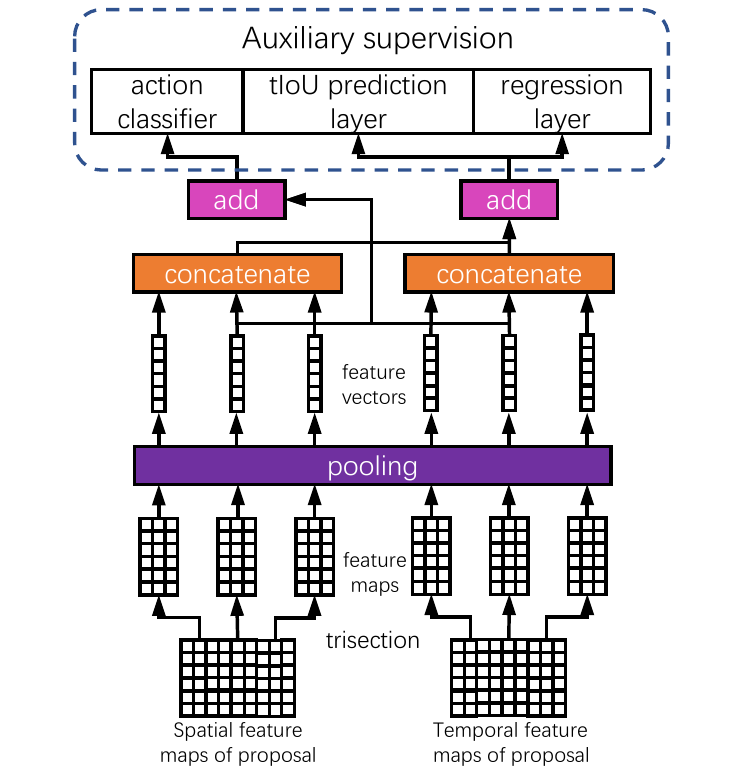}
\end{center}
   \caption{Details of auxiliary supervision. Both spatial and temporal feature maps generated by subnet I are respectively divided into three parts and pooled into three feature vectors. Feature vectors representing middle part of spatial and temporal feature maps are summed and fed into the action classifier. Feature vectors representing each parts of spatial and temporal feature maps are concatenated, summed and fed into the tIoU prediction layer and the regression layer.}
\label{img10}
\end{figure}

\subsection{Auxiliary Supervision}\label{sec4.2}

Auxiliary supervision were added at the end of the subnet I to accelerate the process of training. Specifically, several additional auxiliary loss functions were added to Subnet I. The process of auxiliary supervision is shown in Fig. \ref{img10}.

First, spatial feature maps and temporal feature feature maps extracted by Subnet I were evenly divided into three parts, respectively. Then, average pooling is used to pool each parts of feature map into single vectors. The reason we choose pooling operation as a bridge to connected feature maps and the auxiliary losses is that the pooling operation does not contain any parameters.It allows the gradient signal directly propagate to subnet I, while computation burden increases slightly. Finally, the feature vector representing the middle part of spatial feature map and the feature vector representing the middle part of temporal feature map were added with equal weight and fed into the action classifier. Feature vectors representing each parts of spatial feature maps and temporal feature maps were respectively concatenated, added with equal weight and fed into the tIoU prediction layer and the regression layer.

\subsection{Multi-task Loss} \label{sec4.3}

Positive proposals and background proposals are used to train the activity classifier. The training samples applied to the tIoU prediction layer comprise positive proposals and confusing proposals. However, only positive proposals are used to train the regression FC layer.

We train the Gemini Network by solving a multi-task optimization problem. The principal loss function is the weighted sum of the action category loss (als), the tIoU loss (tIoU), and the regression loss (reg):
\begin{equation}
L_{p}=L_{als}+\lambda\cdot L_{tIoU}+\mu\cdot L_{reg},
\end{equation}
where $\lambda$ and $\mu$ are the weight terms balancing the contribution of each part. The auxiliary loss is as same as the principal loss except the input (as shown in Fig. \ref{img10}). Here, we only use the auxiliary loss in the training phase.

The action category loss $L_{als}$ is used to train an action classifier. We specifically employ the softmax loss over multiple categories. The input proposals are classified into $k+1$ classes, i.e. $K$ activity classes and a background class. In order to avoid the interference of background, the action classifier makes the predictions based on feature $f_j^c$, instead of the total feature  $f_j=[f_j^s,f_j^c,f_j^e]$ . The action category loss can be defined as:
\begin{equation}
L_{als}=\frac{1}{N}\sum_{i=1}^{N}(-log(P_{i}^{(\boldsymbol{a}^{i})})),
\end{equation}
where $P_{i}^{(\boldsymbol{a}_{i})}=\frac{exp(\boldsymbol{p}_{class}^{i})\cdot \boldsymbol{a}^{i}}{\sum_{j}exp(p_{class,j})\cdot a_{j}^{i}}$ ( $\boldsymbol{p}_{class}^i$ is classification score vector of length $k+1$ of the $i$-th proposal, and $\boldsymbol{a}^i$ is the classification label vector of the $i$-th proposal), and $N$ is the number of proposals in each mini-batch.

The tIoU loss $L_{tIoU}$ is used to train the tIoU classifier, which contains $k$ binary classifiers, each for one activity class. We use the total representation $f_j$ obtained from the augmented proposals, which includes the surrounding context. The surrounding context enables the tIoU layer to estimate the degree of overlap between proposal and ground truth. The tIoU loss is given by:
\begin{equation}
L_{tIoU}=\frac{1}{N}\sum_{i=1}^{N}(-log(P_{i}^{(\boldsymbol{c}^{i})})),
\end{equation}
where $P_{i}^{(\boldsymbol{c}_{i})}=\frac{exp(\boldsymbol{p}_{tIoU}^{i})\cdot \boldsymbol{c}^{i}}{\sum_{j}exp(p_{tIoU,j})\cdot c_{j}^{i}}$ ($\boldsymbol{p}_{tIoU}^{i}$ is the tIoU score vector of length $k$ of the $i$-th proposal, and $\boldsymbol{c}^i$ is the tIoU label vector of the $i$-th proposal), and $N$ is the number of proposals in each mini-batch.

The regression loss $L_{reg}$ is a smooth L1 loss function used for regression. The regression loss is given by:
\begin{small}
\begin{equation}
L_{reg}=\frac{1}{N}\sum_{i=1}^{N}({SL}_{1}(\Delta{loc}_{i}'-\Delta{loc}_{i})+{SL}_{1}(\Delta{len}_{i}'-\Delta{len}_{i})),
\end{equation}
\end{small}

\noindent where $SL_1(\cdot)$ is the smooth L1 loss \cite{ren2015faster}, and $\Delta{loc}_{i}'$ and $\Delta{len}_{i}'$ are respectively the predicted relative offsets of location and length. $\Delta{loc}_{i}$ and $\Delta{len}_{i}$ are ground truth relative offsets and are computed as follows:
\begin{equation}
\left\{
             \begin{array}{lr}
             \Delta{loc}_{i}=(loc_{i}^{\ast}-loc_{i})/len_{i} \\
             \Delta{len}_{i}=log(len_{i}^{\ast}/len_{i})
             \end{array}\label{eq7}
\right.
\end{equation}
where $loc_{i}^{\ast}$ and $len_{i}^{\ast}$ respectively denote the center location and span of the $i$-th proposal, and $loc_{i}$ and $len_{i}$ denote those of the corresponding ground truth action segments.

\subsection{Prediction}\label{sec4.4}

The action prediction in this study involved several steps. First, we fed a video into the Gemini Network to generate 1) a set of proposals and 2) three vectors for each proposal, namely, the action category score vector, tIoU score vector, and regression vector. Second, we used non-maximum suppression (NMS) to select some proposals based on their combined scores. Denote $\boldsymbol{a}$ as the action category score, and $\boldsymbol{i}$ as the tIoU score. The combined scores $\boldsymbol{s}$ was computed as follows:
\begin{equation}
\boldsymbol{s}=\boldsymbol{i}\cdot e^{\boldsymbol{a}}
\end{equation}
In practice we found the action category score $\boldsymbol{a}$ to be of higher significance. Therefore we use an exponential function in order to emphasize $\boldsymbol{a}$. We fixed the tIoU threshold of the NMS to 0.2 in THUMOS14, and 0.6 in ActivityNet v1.2. Finally, the boundaries of the selected proposals were further refined based on the regression vectors. The boundary prediction was in the form of the relative displacement of the center point and length of the segments. To determine the start and end times of the predicted proposals, inverse coordinate transformation to Eq. 7 was performed.

\section{Experimental Setting}

\subsection{Datasets}\label{sec5.1}

To evaluate the quality of the  proposed framework, we run experiments (presented in the next section) over two large-scale action detection benchmark datasets, namely, THUMOS14 \cite{THUMOS14} and ActivityNet \cite{Heilbron2015ActivityNet}.

{\bf THUMOS14} \cite{THUMOS14} does not contain untrimmed videos for training. Its official training dataset is UCF101, a trimmed video dataset. UCF101 contains 1010 videos for validation and 1574 videos for testing. Two hundred and twenty of the validation videos and 212 of the testing videos have temporal annotations. In practice, the validation and testing video sets are usually used for training and evaluation, respectively. Two falsely annotated videos (``270'' and ``1496'') in the testing set were excluded in the present study.

{\bf ActivityNet} \cite{Heilbron2015ActivityNet} version 1.2, which contains 9682 videos divided into 100 classes, was used in the present study. The entire dataset is divided into three disjoint subsets, namely, the training, validation, and testing subsets, in the ratio 2:1:1. Compared with THUMOS14, most of the videos in ActivityNet contain activity instances of a single class instead of sparsely distributed actions. The proposed model was trained on the training subset and evaluated on the validation subset.

In our experiments, THUMOS14 was used to analyze the model. Both the THUMOS14 and ActivityNet datasets were used to compare the proposed model with other state-of-the-art methods.

\subsection{Implementation Details}\label{sec5.2}

A pragmatic three-step training algorithm was used to optimize the parameters of the Gemini Network. Subnet I was trained in the first step. The network was initialized using a kinetics-pre-trained model, and then fine-tuned end-to-end for the temporal action localization task. The step-1 loss was the multi-task loss described in Sec. \ref{sec4.3}. In the second step, the parameters of Subnet I were fixed and Subnet II was trained. The initialization of Subnet II was randomly performed. The step-2 loss was the same as the step 1 loss. Finally, in step 3, both Subnet I and Subnet II were fine-tuned. The step-3 loss was the weighted sum of the step-1 and step-2 losses. We consider the losses contribute equally, thus the weights of the step-1 and step-2 losses were set to 0.5 and 0.5, respectively.

The parameters of the proposed framework were learned using SGD, with batch size 128 and momentum 0.9. In each mini-batch, we maintained the ratio of the three types of proposals, namely, positive, confusing, and background proposals, at 1:6:1. For the tIoU prediction layer, only the samples, the loss values of which were ranked in the first 1/6 of a mini-batch, were used to calculate the gradients. To augment the dataset, location jittering, horizontal flipping, corner cropping, and scale jittering techniques were employed. For optical flow extraction, we applied the TVL1 \cite{perez2013tv} optical flow algorithm, which was implemented in OpenCV using CUDA.

\subsection{Evaluation Metrics}\label{sec5.3}

For both datasets, conventional metrics were used to evaluate the average precision (AP) for each action category and calculate the mean average precision (mAP). The prediction of a proposal prediction should be judged to be correct if it meets two requirements: 1) the prediction category is the same as its ground truth; 2) the temporal Intersection-over-Union (tIoU) with its ground truth instance is greater than the tIoU threshold $\theta$. 

For the dataset ActivityNet v1.2, the tIoU thresholds were \{0.5, 0.75, 0.95\}. For the dataset THUMOS14, the tIoU thresholds were \{0.1, 0.2, 0.3, 0.4, 0.5, 0.6, 0.7\}. In addition, the average mAPs was calculated under tIoU thresholds \{0.5,0.55,0.6...0.9,0.95\} for ActivityNet v1.2.

\section{Experimental Results}

The proposed framework was applied to two test datasets, described in Sec. \ref{sec5.1}. We first compares its performance with those of other state-of-the-art approaches, and then presents qualitative visualization and failure cases analysis, respectively. Finally, we presents the analysis of different modules of the Gemini Network.

\subsection{Comparison with State-of-the-art Methods}\label{sec6.1}

\begin{figure*}[t]
\begin{center}
% \fbox{\rule{0pt}{2in} \rule{.9\linewidth}{0pt}}
\includegraphics[width=16.5cm]{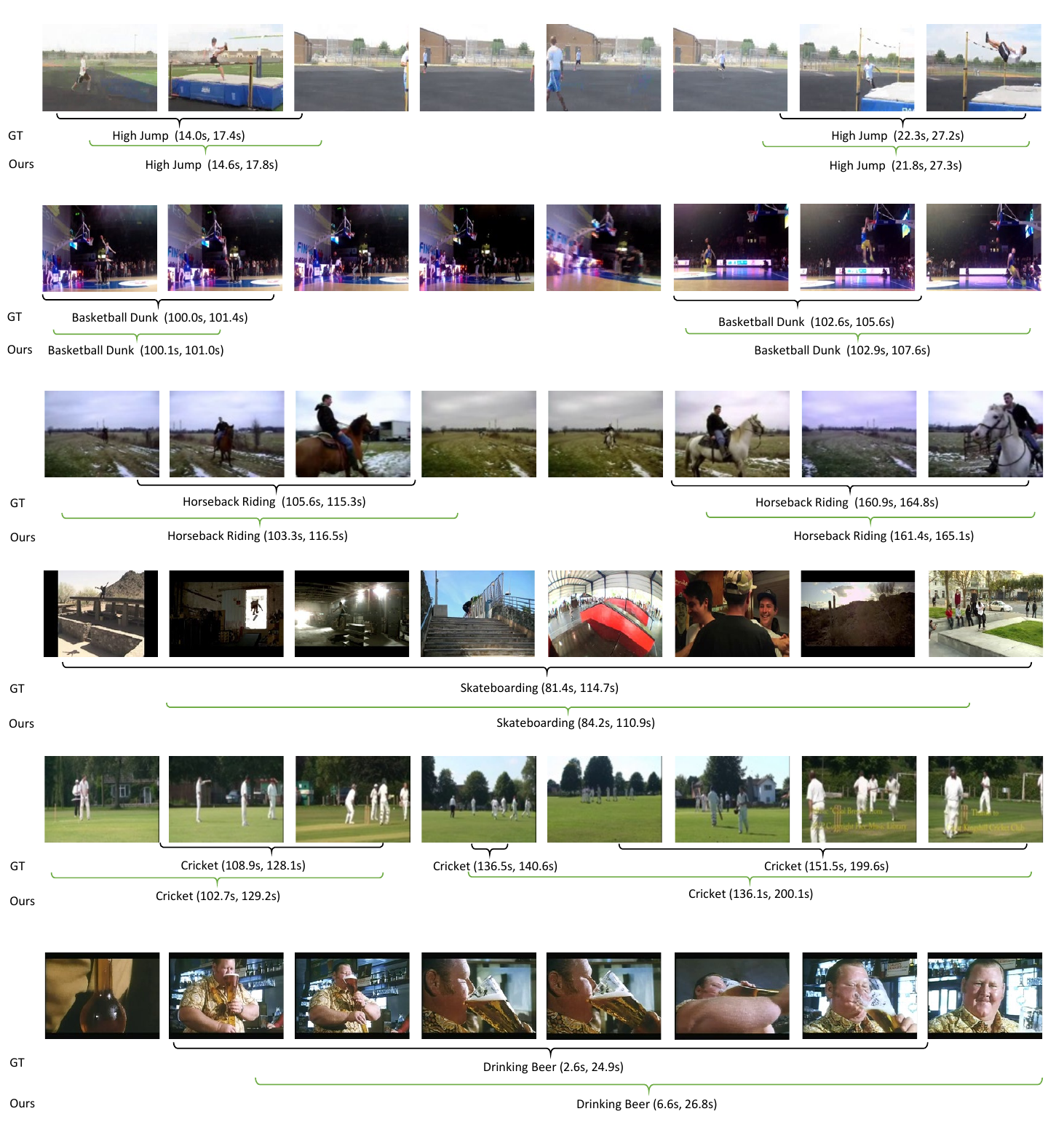}
\end{center}
   \caption{Qualitative visualization of activities predicted by the Genimi Network. The videos are shown as sequences of frames, where GT indicates ground truth. The ground truth activity segments are marked in black, while the predicted activity segments are marked in green for correct prediction on the condition tIoU$\geqslant0.5$.}
\label{img7}
\end{figure*}

\begin{figure*}[t]
\begin{center}
% \fbox{\rule{0pt}{2in} \rule{.9\linewidth}{0pt}}
\includegraphics[width=16.5cm]{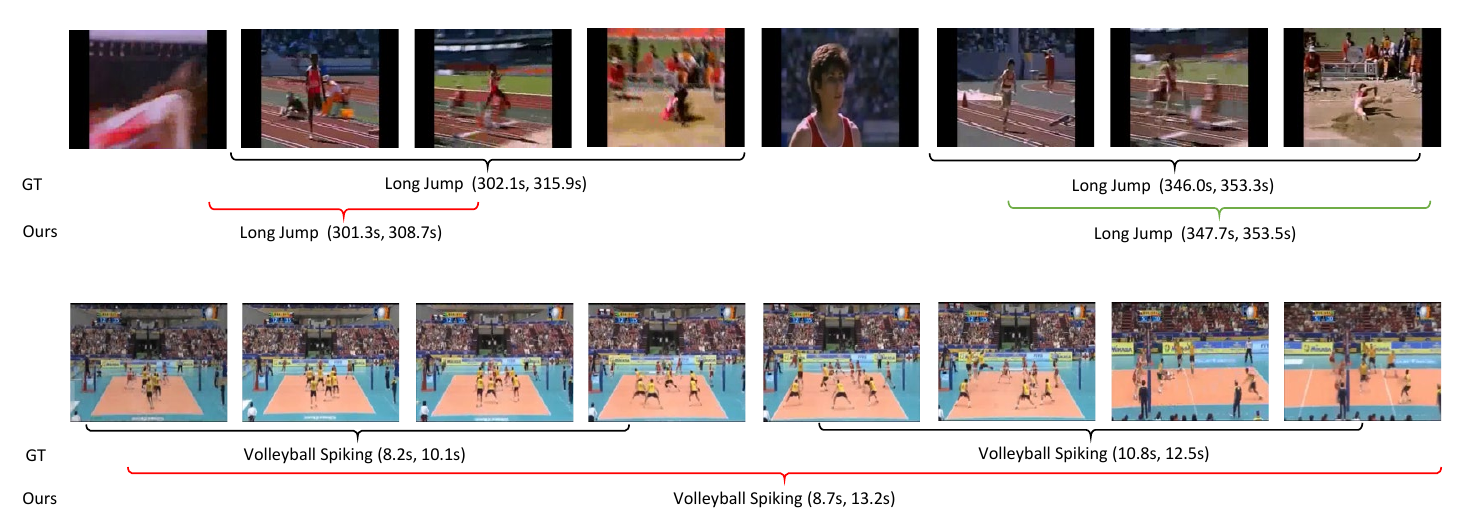}
\end{center}
   \caption{Failure cases. The ground truth activity segments are marked in black, while the predicted activity segments are marked in red for wrong prediction on the condition tIoU$\leqslant0.5$.}
\label{img8}
\end{figure*}

We compared the proposed method with other state-of-the-art temporal action localization methods using the THUMOS14 \cite{THUMOS14} and ActivityNet v1.2 \cite{Heilbron2015ActivityNet} datasets, based on the above-mentioned performance metrics. The average action duration and the average video duration in THUMOS14 were observed to be 4 and 233 s, respectively, while those in ActivityNet v1.2 were 50 and 114 s. This indicates significant difference between the temporal structures and granularities of the two datasets. The proposed framework would thus have to be highly adaptable to perform consistently well on both datasets.

\begin{table}[t]
\caption{Temporal action localization result on THUMOS14, measured by mAP at different tIoU thresholds $\theta$.}
\begin{center}
\resizebox{0.47\textwidth}{!}{
\begin{tabular}{c|ccccccc}
  \hline
  \multicolumn{8}{c}{THUMOS14, mAP(\%)@$\theta$} \\ \hline
  tIoU & 0.1 & 0.2 & 0.3 & 0.4 & 0.5 & 0.6 & 0.7 \\ \hline
  Shou et al. \cite{shou2017cdc} & - & - & 40.1 & 29.4 & 23.3 & 13.1 & 7.9 \\
  Yuan et al. \cite{yuan2017temporal} & 51.0 & 45.2 & 36.5 & 27.8 & 17.8 & - & - \\
  Buch et al. \cite{buch2017end} & - & - & 45.7 & - & 29.2 & - & 9.6 \\
  Gao et al. \cite{Gao2017Cascaded} & 60.1 & 56.7 & 50.1 & 41.3 & 31.0 & 19.1 & 9.9 \\
  Hou et al. \cite{hou2017real} & 51.3 & - & 43.7 & - & 22.0 & - & - \\
  Dai et al. \cite{dai2017temporal} & - & - & - & 33.3 & 25.6 & 15.9 & 9.0 \\
  Gao et al. \cite{gao2017turn} & 54.0 & 50.9 & 44.1 & 34.9 & 25.6 & - & - \\
  Xu et al. \cite{xu2017r} & 54.5 & 51.5 & 44.8 & 35.6 & 28.9 & - & - \\
  Nguyen et al. \cite{nguyen2018weakly} & 52.0 & 44.7 & 35.5 & 25.8 & 16.9 & 9.9 & 4.3 \\
  Paul et al. \cite{paul2018w} & 55.2 & 49.6 & 40.1 & 31.1 & 22.8 & - & 7.6 \\
  Alwassel et al. \cite{alwassel2018action} & - & - & 51.8 & 42.4 & 30.8 & 20.2 & 11.1 \\
  Chao et al. \cite{chao2018rethinking} & 59.8 & 57.1 & 53.2 & 48.5 & \textbf{42.8} & \textbf{33.8} & 20.8 \\
  Song et al. \cite{8440749} & 48.5 & 44.1 & 38.2 & 29.8 & 20.1 & - & - \\
  Li et al. \cite{8476583} & 61.2 & 57.3 & 50.6 & 41.0 & 30.8 & 19.3 & 8.6 \\
  Wang et al. \cite{8579103} & 47.4 & 45.9 & 39.4 & 33.0 & 26.5 & - & - \\
  \hline
  Ours & \textbf{63.5} & \textbf{61.0} & \textbf{56.7} & \textbf{50.6} & 42.6 & 32.5 & \textbf{21.4} \\ \hline
\end{tabular}}
\end{center}
\label{tab6}
\end{table}

\begin{table}[t]
\newcommand{\tabincell}[2]{\begin{tabular}{@{}#1@{}}#2\end{tabular}}
\caption{Temporal action localization result on ActivityNet v1.2, measured by mean average precision (mAP) at different tIoU thresholds $\theta$ and the average mAP of tIoU thresholds from 0.5 to 0.95.}
\begin{center}
\resizebox{0.47\textwidth}{!}{
\begin{tabular}{c|c|ccc}
  \hline
  \multicolumn{5}{c}{ActivityNet v1.2, mAP(\%)@$\theta$} \\ \hline
  \multirow{2}{*}{tIoU} & \multirow{2}{*}{\tabincell{c}{Average of\\\{0.5,0.55,...,0.95\}}} & \multicolumn{3}{c}{Some examples} \\ \cline{3-5}
  ~ & ~ & 0.5 & 0.75 & 0.95 \\ \hline
  Singh and Cuzzolin \cite{singh2016untrimmed} & 11.3 & 22.7 & 10.8 & 0.3 \\
  Singh \cite{singh2016action} & 14.6 & 26.0 & 15.2 & 2.6 \\
  Wang and Tao \cite{wang2016uts} & 16.4 & 45.1 & 4.1 & 0.0 \\
  Yang et al. \cite{yang2018one} & 10.0 & 23.1 & - & - \\
  Shou et al. \cite{shou2017cdc} & 23.8 & 45.3 & 26.0 & 0.2 \\
  Paul et al. \cite{paul2018w} & 18.0 & 37.0 & - & - \\
  Yuan et al. \cite{yuan2017temporal} & 24.9 & 41.1 & 24.6 & 5.0  \\
  Li et al. \cite{8476583} & 25.0 & 42.3 & 24.9 & 5.2  \\
  \hline
  Ours & \textbf{33.6} & \textbf{50.4} & \textbf{34.9} & \textbf{8.0}  \\
  \hline
\end{tabular}}
\end{center}
\label{tab7}
\end{table}
\noindent{\bf Results on THUMOS14:} We compared the proposed method with other state-of-the-art temporal action localization methods using the THUMOS14 \cite{THUMOS14} based on the above-mentioned performance metrics. The results of the comparison of the present model with existing state-of-the-art methods on THUMOS14 are shown in Table \ref{tab6}. It can be seen that, in most cases, the proposed method outperforms the existing methods. Figure \ref{img4}(a) shows some representative qualitative results in both datasets.

\noindent{\bf Results on ActivityNet:} The results of the comparison for ActivityNet v1.2 are presented in Table \ref{tab7}. As references, the table also gives the performances of some recent works. The proposed framework achieved an average mAP of 33.6, which clearly demonstrates its superiority.

\subsection{Qualitative Visualization}\label{sec6.2}
Besides localization accuracies, we would like to attain further insight into the difference between predicted action instance and ground truth. We pick two classes from the THUMOS14 dataset (i.e., High Jump and Basketball Dunk), and four classes from the ActivityNet dataset (i.e., Horseback Riding, Skateboarding, Cricket, and Drinking Beer). The results are illustrated in Fig. \ref{img7}. The videos are shown as sequences of frames, where GT indicates ground truth. The ground truth activity segments are marked in black, while the predicted activity segments are marked in green for correct prediction on the condition tIoU$\geqslant0.5$. As shown in Fig. \ref{img7}, the proposed framework is highly adaptable to all kinds of actions and performs consistently well on both datasets.

\subsection{Failure Cases}\label{sec6.3}
In most cases, the Gemini Network can detect actions very well. However, we would like to show some failure cases here in Fig. \ref{img8}. The segments marked in red are the wrong ones. In the volleyball spiking video, the duration of background between two volleyball spiking instances is much shorter than the duration of each volleyball spiking instance (0.7s vs 1.9s, 0.7s vs 1.7s). In such a situation, Gemini Network may wrongly regard two instances as single one. For another example, in the long jump video, the drastic camera motion makes the Gemini Network find a wrong end point of the first long jump instance.

\subsection{Model Analysis}\label{sec6.4}

\begin{figure}[t]
\begin{center}
\includegraphics[width=8.2cm]{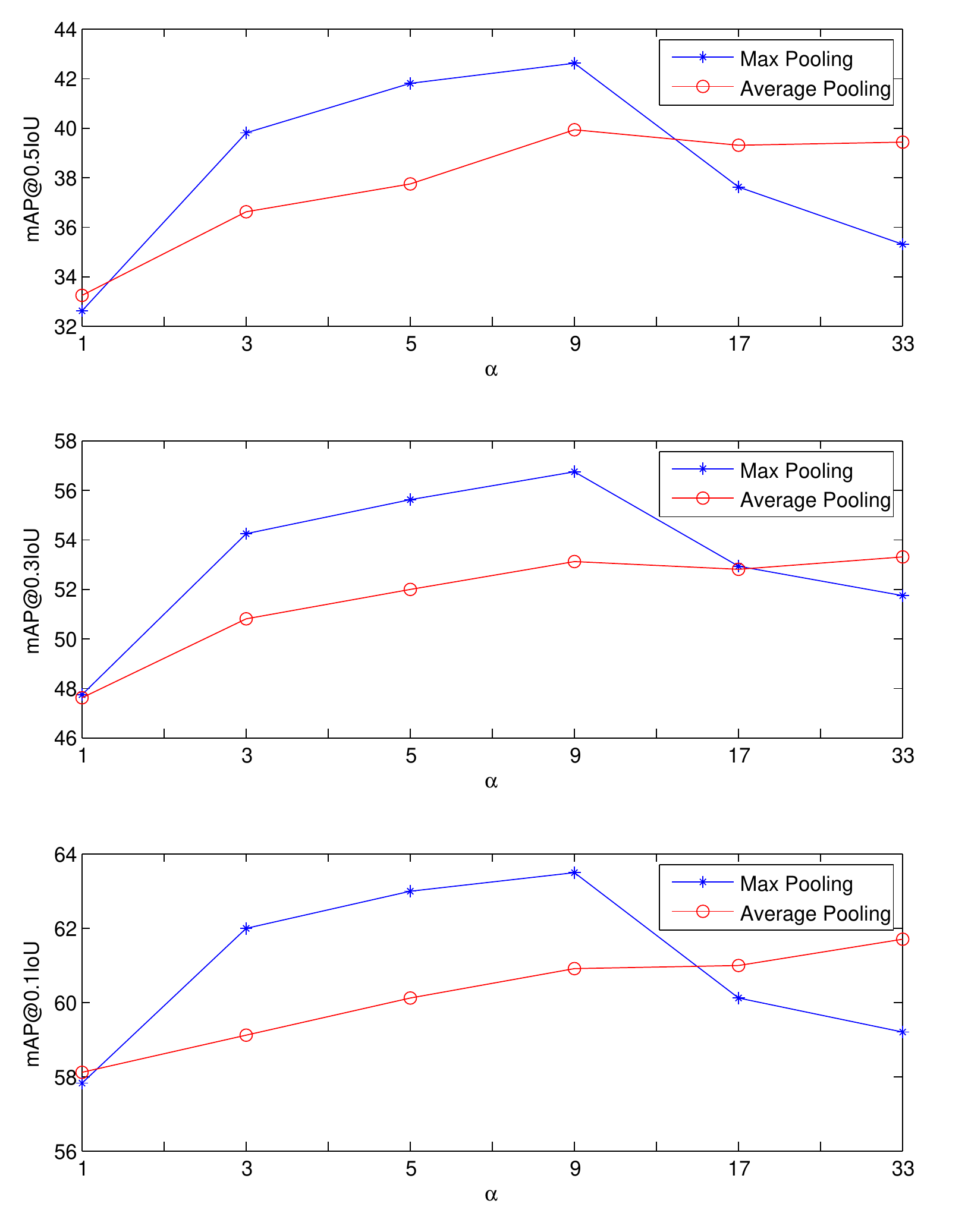}
\end{center}
   \caption{Comparison between different $\alpha$ and pooling methods. Results are measured by mAP at tIoU threshold 0.5(top), 0.3(middle) and 0.1(bottom). $\alpha$ indicates the number of segments.}
\label{img5}
\end{figure}

\noindent{\bf Comparison of Different Segment Numbers.}  We attempted to determine the most appropriate number of segments to be used for each strategy. For this purpose, in the case of THUMOS14, the number of segments, $\alpha$, was varied as 1, 3, 5, 9, 17, and 33, respectively. We configured the dimension of feature vector $d_j$ as 129. The results are summarized in Fig. \ref{img5}. We observed that increasing the number of segments, $\alpha$, initially increased the performance significantly. However, when $\alpha$ exceeded a particular threshold, the performance no longer increased, and actually decreased in the case of max pooling.
\\

\begin{figure}[t]
\begin{center}
\includegraphics[width=8.5cm]{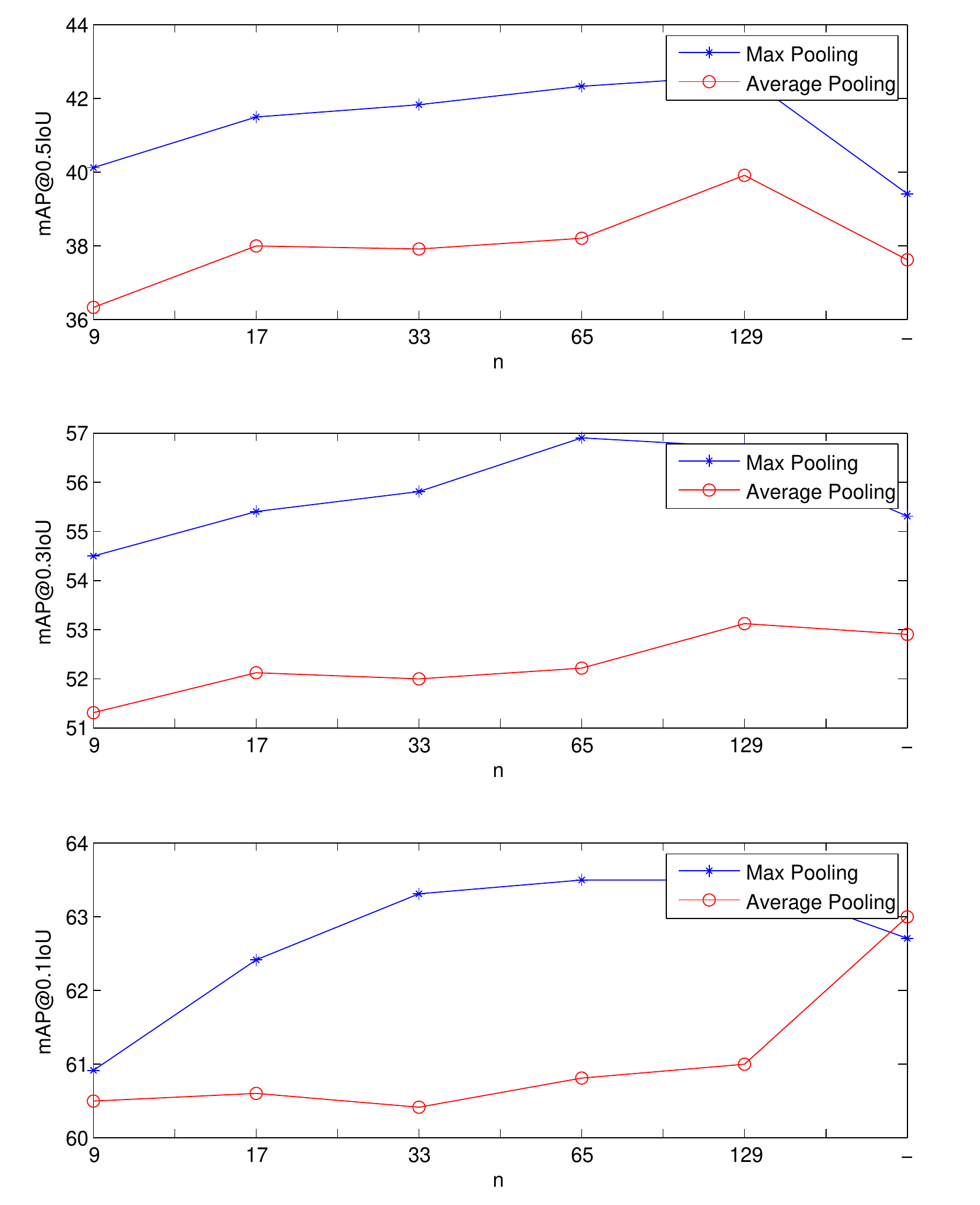}
\end{center}
   \caption{Comparison between different $n$ and pooling methods. Results are measured by mAP at tIoU threshold 0.5(top), 0.3(middle) and 0.1(bottom). $n$ indicates the dimension of unit-level feature vector.}
\label{img6}
\end{figure}

\noindent{\bf Comparison of Feature Recoding.}  We also investigated the influence of the dimension $n$ of the feature vector during feature recoding. For this purpose, to control the computational complexity and performance, the number of segments, $\alpha$, was set to 9 and the feature vector dimension $n$ was varied as 9, 17, 33, 65, and 129. We attempted to determine the most appropriate value of $n$, as well as verify the importance of the recoding function by comparing the results with that of a model that ablates the function. From the results summarized in Fig. \ref{img6}, it can be observed that the performance without feature recoding is always inferior, irrespective of the pooling method. This demonstrates the importance of feature recoding for accurate localization. The performance is also unsatisfactory when the dimension $n$ is less than 65. This is probably due to the recoded feature vector losing some information useful for temporal action localization when the dimension is low.
\\

\noindent{\bf Comparison of Different Temporal Pooling Strategies.}  Here, we investigated the influence of the different temporal pooling strategies. For this purpose, in the case of THUMOS14, we configured the dimension of feature vector $d_j$ as 129 for the comparison of two temporal pooling methods, namely, average pooling and max pooling. The results are summarized in Fig. \ref{img5}. It can be observed from the results in Fig. \ref{img5} that max pooling is more suitable for structured temporal pooling when the number of segments is small.
\\

\noindent{\bf Comparison of Temporal Structure Modeling.}  Here, we compared different methods for long-range temporal structure modeling, namely, temporal average pooling, temporal convolution, and 2D convolution. In addition, different convolutional filters were tested to determine the most suitable architecture. From the results presented in Table \ref{tab4}, it can be seen that 2D convolution is better for modeling a long-range temporal structure compared with temporal average pooling and temporal convolution. We also compared the performances of the proposed network when Subnet II is removed with when both subnets are used. It was found that the use of only one subnet to model the temporal structure diminished the performance. This result demonstrates the importance of Subnet II to long-range temporal structure modeling.
\\

\noindent{\bf Comparison of Action Classifier, tIoU Prediction Layer and Location Regression Layer.}  Here, we verify the importance of tIoU prediction layer by comparing another design that uses only the activity classifiers to classify the proposals. From the results shown in Table \ref{tab5}, it can be observed that the use of only action classifier to distinguish between the positive proposals and the different types of negative proposals diminished the performance, with mAP decreasing from 37.5\% to 33.5\%. The previous higher performance is attributed to the use of the tIoU prediction layer, which subdivides the negative samples and improves the discrimination ability of the classifier.
\begin{table}[t]
\caption{Comparison between different methods for long-range temporal structure modeling . Results are measured by mAP at tIoU threshold 0.5}
\begin{center}
\resizebox{0.47\textwidth}{!}{
\begin{tabular}{c|c|c}
  \hline
  Method & Kernel Size & mAP(\%) \\ \hline
  Temporal Avg Pool & - & 30.6 \\ \hline
  \multirow{2}{*}{Temporal Conv} & 3 & 35.9 \\ \cline{2-3}
  ~ & 5 & 35.6 \\ \hline
  \multirow{2}{*}{1D Conv + 1D Conv} & $1\times 3\ \&\ 3\times 1$ & 39.2 \\ \cline{2-3}
  ~ & $1\times 5\ \&\ 5\times 1$ & 39.4 \\ \hline
  \multirow{2}{*}{2D Conv} & $3\times 3$ & 39.9 \\ \cline{2-3}
  ~ & $5\times 5$ & 39.3 \\ \hline
\end{tabular}}
\end{center}
\label{tab4}
\end{table}

\begin{table}[h]
\caption{Ablation study on THUMOS14. Act. refers to action classifiers and tIoU Reg. refers to tIoU prediction layer. Loc. Reg. denotes the use of location regression. Results are measured by mAP at tIoU threshold 0.5.}
\begin{center}
\resizebox{0.47\textwidth}{!}{
\begin{tabular}{c|cccc|cc}
  \hline
  Training&\multicolumn{4}{c|}{1st + 2nd}&\multicolumn{2}{c}{1st + 2nd + 3rd} \\ \hline
  Subnet I&$\surd$&$\surd$&$\surd$&$\surd$&$\surd$&$\surd$ \\
  Subnet II& &$\surd$&$\surd$&$\surd$&$\surd$&$\surd$ \\
  Act.&$\surd$&$\surd$&$\surd$&$\surd$&$\surd$&$\surd$ \\
  tIoU Pre. & & &$\surd$&$\surd$&$\surd$&$\surd$ \\
  Loc. Reg. & & & &$\surd$& &$\surd$ \\ \hline
  mAP&28.2&33.5&37.5&42.2&37.7&42.6 \\ \hline
\end{tabular}}
\end{center}
\label{tab5}
\end{table}

Because the augmented proposals contain the entire contextual information of the action instances, the starting and ending points of the proposals can be precisely regressed. The contribution of the location regression to the localization performance was evaluated, as detailed in Table \ref{tab5}. From the results, it can be seen that the location regression prominently boosts the performance. For further investigation, we eliminated the third step in the training of the present model. The obtained results are presented in Table \ref{tab5}, from which it can be observed that the third-step training improves the performance. The results demonstrate the superiority of joint fine-tuning networks.

\section{Conclusion}
In this study, we developed a generic framework with two subnets, referred to as Gemini Network, for temporal action localization. By dividing the temporal structure into two types of structures and respectively modeling the latter in the two subnets, we achieved significant performance improvement compared with existing state-of-the-art methods. We also demonstrated that the use of auxiliary supervision, structured temporal pooling, and feature recoding enhanced the performance of the framework.

%\section*{Acknowledgment}

\ifCLASSOPTIONcaptionsoff
  \newpage
\fi



%\bibliographystyle{IEEEtran}

\bibliographystyle{IEEEtran}
\bibliography{egbib}


%\end{IEEEbiography}
%% insert where needed to balance the two columns on the last page with
%%\newpage


%\vfill


\end{document}